# Skin Lesion Classification using Class Activation Map

Xi Jia and Linlin Shen

Computer Vision institute, Shenzhen University

**Abstract -** We proposed a two stage framework with only one network to analyze skin lesion images, we firstly trained a convolutional network to classify these images, and cropped the import regions which the network has the maximum activation value. In the second stage, we retrained this CNN with the image regions extracted from stage one and output the final probabilities. The two stage framework achieved a mean AUC of 0.857 in ISIC-2017 skin lesion validation set and is 0.04 higher than that of the original inputs, 0.821.

## 1. INTRODUCTION

Deep convolutional neural networks have been proved to be efficient in many fields of computer vision, such as natural image classification[1], face recognition[2] and so on. There are many literatures use CNN methods to analyze skin lesions as well. Codella et.al [3] used a pre-trained convolutional network as a feature extractor to classify skin lesions and their experiments showed that the deep features outperform traditional low-level visual features. In order to get multiscale features, Kawahara et.al [4] converted AlexNet into a Fully Convolutional network by converting the fully connected layers into convolutional layers, then a logistic classifier was trained to classify the skin lesions.

Zhou et.al[5] has shown that there is no location information when training a classification network, convolutional neural networks still have remarkable abilities for localizing objects. They proposed a technique called Class Activation Mapping (CAM) by using global average pooling (GAP) in CNNs. A class activation map for a particular category indicates the discriminative image regions used by the CNN to identify the category.

In this work, we adopted the CAM technique to help us analyze skin lesions. We find that the import regions of CAM brings a 0.04 gain of the mean AUC.

## 2. SYSTEM OVERVIEW

*2.1 The Flow chart of system*

As shown in Figure 1, we input a skin lesion image into CNN and get its class activation maps in stage 1. And then we crop the import regions from original images according to CAM and use them as the input of stage 2 to produce the final probabilities.

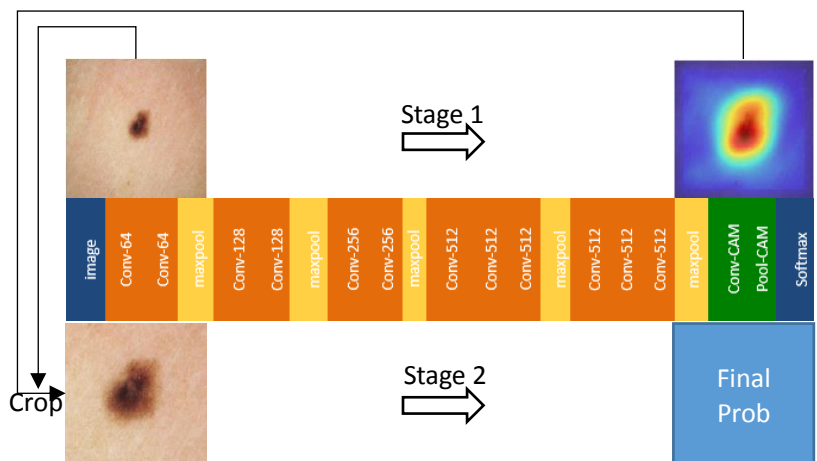

Figure 1, System overview. Get the CAM (heatmap on the upper right corner) of input image in stage 1. Crop the import regions as input of stage 2 and get the final score.

*2.2 Network Architecture*

We used a deep convolutional neural network with 14 convolutional layers and one fully connected layer to classify and analyze skin lesion images. This network is the same as VGG-GAP[5] except there is no padding in last two convolutional layers.

There were five pooling layers in the whole network, and the first four of them used 2*2 max pooling with strides 2. We used Global Average Pooling structure in the last pooling layer since it acts as a structural regularizer, and retain CNN's localization ability until the final layer[5, 6]. We used 0.5 dropout followed by GAP layer to reduce overfitting in the model training procedures. In the last FC layer, we also used softmax to output the probability of each categories as a classification function.

*2.3 Class Activation Map*

In a convolutional neural network, for a given class $c$, if we use $S_c$ denotes the input of the softmax layer, then the output of the softmax can be computed by $\frac{\exp(S_c)}{\sum \exp(S_c)}$. The formula of $S_c$ as shown in equation 1.

$$S_c = \sum_k w_k^c \sum_{x,y} f_k(x,y) = \sum_{x,y} \sum_k w_k^c f_k(x,y) \qquad eq.1$$

Where $f_k(x, y)$ represents the activation value of unit $k$ in the last convolutional layer at spatial location $(x, y)$, $w$ indicates the importance of $f_k$ for class $c$.

If we use back-propagation to get the value of $w$ when network training were done, then the CAM of class $c$, denoted as $Mc$, can be computed by equation 2. Figure 2 shows an input skin lesion image and its CAMs corresponding to the three classes.

$$M_c = \sum_k w_k^c f_k(x,y) \qquad eq.2$$

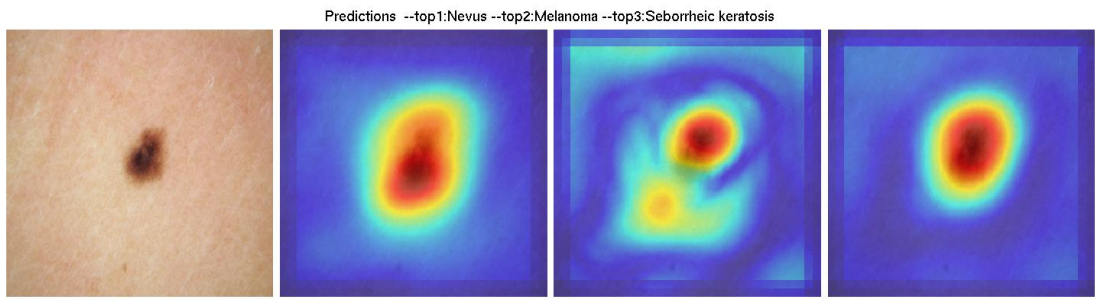

Figure 2. Examples of a skin lesion image and its three CAMs (from left to right), the ground truth of this image is Nevus.

### 3. DATASETS AND PRE-PROCESSING

*3.1 Datasets Descriptions*

There are 2000 images in ISIC-2017 training sets, including 374 "melanoma", 254 "seborrheic keratosis", and the remainder as benign nevi (1372). A separate validation dataset (about 150 images) was used for participants to submit automated results for evaluation. The final test dataset contains 600 images, which were remained by the organizers to decide the final winner of this competition. The medical descriptions of the three skin lesions were described as follows:

    Melanoma – malignant skin tumor, derived from melanocytes (melanocytic)
    Nevus – benign skin tumor, derived from melanocytes (melanocytic)
    Seborrheic keratosis – benign skin tumor, derived from keratinocytes (non-melanocytic)

*3.2 Preprocessing*

Before feeding the images into our CNN architecture, we used a popular method proposed by Lee et.al [7] to remove hairs. This method consists of following steps, firstly it use a generalized grayscale morphological closing operation to locate the dark hair, and once the dark hair pixels were verified by thin and long structure, the hair pixels will be replaced by a bilinear interpolation and then smoothed by an adaptive median filter.

*3.3 Data Augmentation*

Since the training set only contains 2000 images, so we augmented these images by rotations . We rotated each image with a 90 degree interval and the amounts of training images were increased to 6000. Additionally, we horizontal flipped the images and cropped five patches from each image.

4 EXPERIMENTS AND RESULTS

*4.1 fine-tuning*

As we mentioned before, the amounts of training images is insufficient for training a deep convolutional network from scratch, so we used the pre-trained VGG-GAP model as initialization of network parameters and fine-tuned the last serval layers. Fine-tuning makes the network much easier to be trained and prevents it from overfitting. We plotted the training curves in figure 3. The left one is training this network from scratch, the right one is fine-tuning, obviously, fine-tuned network can lead to a better convergence. In our experiments, we used Caffe[8], a deep learning library to train our models.

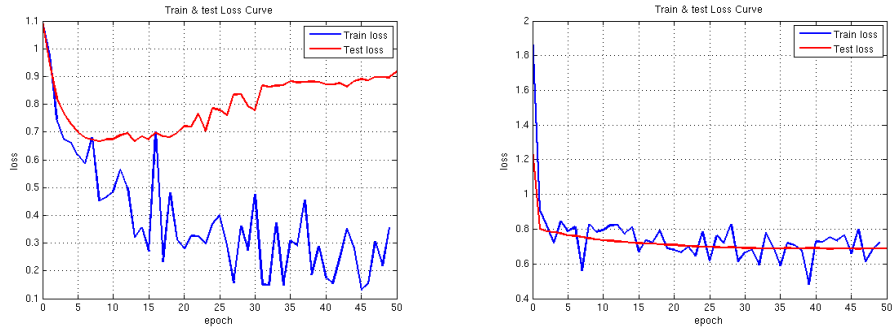

Figure 3. The upper one is loss curve training this network from scratch, another one is fine-tuning.

*4.2 Visualization*

Since the scale of skin lesion images are usually large and involve many irrelevant regions, in our experiments we found sometimes the CNN could be confused and cannot get the right regions as shown in Figure 4.

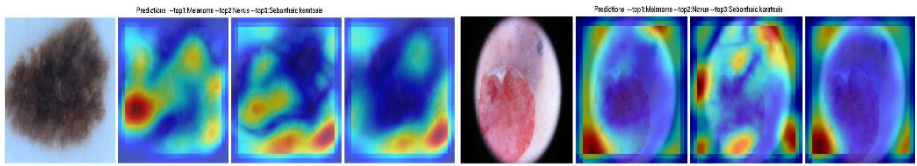

Figure 4. Two examples of CNN made errors about import regions.

*4.3 Results*

We listed serval results in table 1, we can observe that after stage1 and stage2, the M_AUC, SK_AUC and AVG_AUC are all higher than stage1.

Table 1. AUC of Stage1 and Stage1,2

| Methods | M_AUC | SK_AUC | AVG_AUC |
|---|---|---|---|
| Stage1 | 0.782 | 0.860 | 0.821 |
| Stage1,2 | 0.807 | 0.909 | 0.858 |